\documentclass[10pt,twocolumn,letterpaper]{article}

\usepackage[pagenumbers]{iccv} 

\usepackage[accsupp]{axessibility}

\definecolor{iccvblue}{rgb}{0.21,0.49,0.74}
\usepackage[pagebackref,breaklinks,colorlinks,allcolors=iccvblue]{hyperref}

\usepackage{multirow}
\usepackage{colortbl}
\usepackage{mathptmx}
\usepackage[T1]{fontenc}
\usepackage{pslatex}
\pdfmapfile{+sansmathaccent.map}
\pdfmapfile{+texnansi.map}

\def\paperID{1322} 
\def\confName{ICCV}
\def\confYear{2025}

\title{Global Regulation and Excitation via Attention Tuning for Stereo Matching}

\author{
Jiahao LI$^{1}$ \quad Xinhong Chen$^{1}$ \quad Zhengmin JIANG$^{1}$ \quad Qian Zhou$^{1}$ \quad Yung-Hui Li$^{2}$ \quad Jianping Wang$^{1}$ \\
$^{1}$City University of Hong Kong $\quad$ $^{2}$Hon Hai Research Institute \\
{\tt\small jiahali2-c@my.cityu.edu.hk}
}

\begin{document}
\maketitle
\begin{abstract}
Stereo matching achieves significant progress with iterative algorithms like RAFT-Stereo and IGEV-Stereo. However, these methods struggle in ill-posed regions with occlusions, textureless, or repetitive patterns, due to a lack of global context and geometric information for effective iterative refinement. To enable the existing iterative approaches to incorporate global context, we propose the \textbf{G}lobal \textbf{R}egulation and \textbf{E}xcitation via \textbf{A}ttention \textbf{T}uning (GREAT) framework which encompasses three attention modules. Specifically, Spatial Attention (SA) captures the global context within the spatial dimension, Matching Attention (MA) extracts global context along epipolar lines, and Volume Attention (VA) works in conjunction with SA and MA to construct a more robust cost-volume excited by global context and geometric details. To verify the universality and effectiveness of this framework, we integrate it into several representative iterative stereo-matching methods and validate it through extensive experiments, collectively denoted as GREAT-Stereo. This framework demonstrates superior performance in challenging ill-posed regions. Applied to IGEV-Stereo, among all published methods, our GREAT-IGEV ranks first on the Scene Flow test set, KITTI 2015, and ETH3D leaderboards, and achieves second on the Middlebury benchmark. Code is available at \href{https://github.com/JarvisLee0423/GREAT-Stereo}{https://github.com/JarvisLee0423/GREAT-Stereo}.
\end{abstract}
    
\section{Introduction}
\label{sec:intro}
\begin{figure}[tp]
    \centering
    \includegraphics[width=0.97\linewidth]{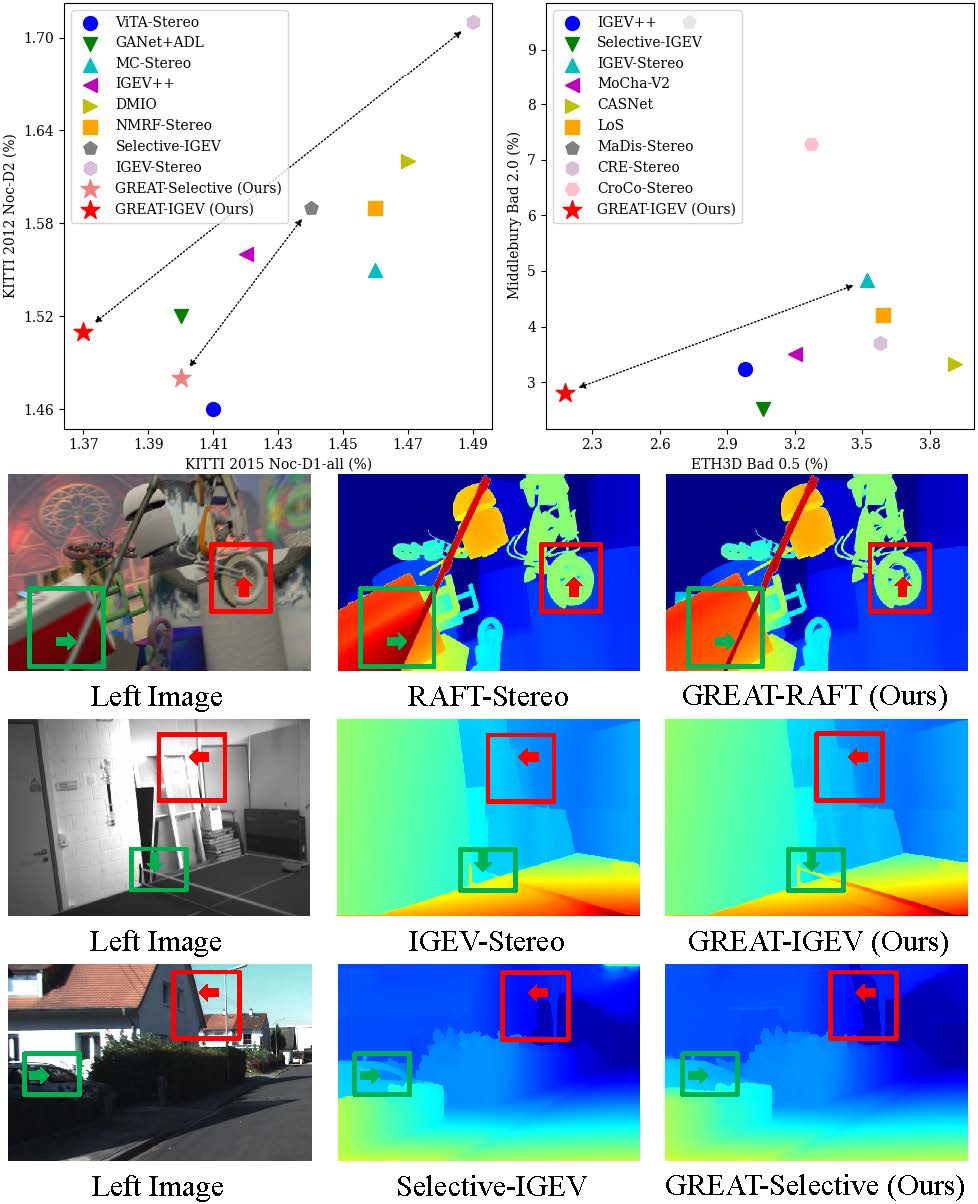}
    \caption{\textbf{Row 1:} Comparisons with state-of-the-art stereo methods on KITTI 2012 \cite{geiger2012we}, KITTI 2015 \cite{menze2015object}, ETH3D \cite{schops2017multi}, and Middlebury \cite{scharstein2014high} leaderboards, where the lower metrics indicate better performance. \textbf{Row 2:} Visual comparison with RAFT-Stereo on Scene Flow \cite{mayer2016large}. \textbf{Row 3:} Visual comparison with IGEV-Stereo on ETH3D. \textbf{Row 4:} Visual comparison with Selective-IGEV on KITTI 2012. Our models produce clearer and more consistent geometric structures in ill-posed regions (green and red boxes).}
    \label{fig:performance_disp_vis}
\end{figure}

Recovering depth information from 2D images is vital in 3D scene reconstruction which improves scene perception in fields like robotics and autonomous driving. Stereo matching intends to predict dense 3D representations referred to as disparity. This technique constructs the pixel-wise cost volume from rectified stereo images to depict matching similarity along epipolar lines and regresses disparity by refining cost volume \cite{hirschmuller2005accurate}.

Taking advantage of deep learning, the current state-of-the-art stereo-matching algorithms are primarily categorized as aggregation-based and iteration-based approaches. Aggregation-based methods \cite{chang2018pyramid, guo2019group, xu2022attention, xu2023accurate, wang2024cost, wang2025adstereo, wang2025dualnet, kendall2017end} leverage convolutional neural network (CNN) to aggregate geometric structures within the pixel-wise cost volume to predict disparity. While these methods achieve superior accuracy, the substantial computational cost makes them impractical for high-resolution images. To address this limitation, researchers turn to iterative schemes \cite{lipson2021raft, li2022practical, zeng2023parameterized}, which avoid computationally expensive cost-volume aggregation. Specifically, these methods employ recurrent units \cite{lipson2021raft} to progressively update the disparity based on local information retrieved from the pixel-wise cost volume, allowing for high-resolution inference.

Despite these advancements, iterative methods still encounter challenges in ambiguous ill-posed regions (occluded, textureless, and repetitive texture areas) due to merely considering pixel-wise and local contextual details. Specifically, the CNN feature extractors widely adopted by iterative approaches use a fixed receptive field size, restricting each pixel feature to encode only local context with its nearest neighboring pixels. Meanwhile, iterative methods rely on a cost-volume constructed on a pixel-by-pixel basis to compute matching similarity \cite{lipson2021raft, zhao2023high, zeng2023parameterized}, thereby the points in ill-posed regions like textureless and repetitive areas with similar local features cannot be effectively distinguished and matched \cite{xu2023iterative}. Moreover, the recurrent units adopted by iterative schemes cannot effectively refine the geometry in disparity based on limited local information in each pixel \cite{wang2024selective}, resulting in blurry geometric structures of different objects and incorrect matching between images.

Based on the above observations, the key to improving the stereo-matching performance of iterative methods in ill-posed regions is to introduce global context information. Specifically, for occluded areas, the complementary information from the global context beyond local neighborhoods can help propagate the geometric structures from non-occluded to occluded regions and yield reliable predictions in occlusions, as demonstrated in GMA-Flow \cite{jiang2021learning}. As for textureless and repetitive texture regions, although these areas are challenging to distinguish at the pixel or local level due to similar patterns, assigning them to the correct geometric structures becomes feasible when the receptive fields are expanded to encompass the global context. Therefore, in this work, we propose to enable the existing iterative methods to incorporate global context details so that they can robustly handle the matching ambiguities in ill-posed regions.

To this end, we introduce a universal framework called \textbf{G}lobal \textbf{R}egulation and \textbf{E}xcitation via \textbf{A}ttention \textbf{T}uning (GREAT) that contains three attention modules. \textbf{Spatial Attention} (SA) encodes global context in spatial dimension from local to global into each pixel, accelerating the propagation of geometric structures within the cost volume. \textbf{Matching Attention} (MA) aggregates global context for each pixel along epipolar lines, effectively reducing ambiguities in pixel pair matching. \textbf{Volume Attention} (VA) excites global context in specific regions of the cost volume by incorporating SA and MA, enhancing its robustness. Based on these modules, as illustrated in Fig. \ref{fig:performance_disp_vis}, the proposed framework—\textbf{GREAT-Stereo} produces clearer and more consistent geometric structures in ill-posed regions compared to baseline by leveraging global context details, thereby enhancing the accuracy in such regions. We validate the effectiveness of our framework in several experiments. When applied to IGEV-Stereo \cite{xu2023iterative}, on the Scene Flow dataset, our GREAT-IGEV achieves the new state-of-the-art overall EPE of 0.41 and non-occluded EPE of 0.14. Compared to the state-of-the-art occluded EPE of 1.53 in GOAT-Stereo \cite{liu2024global}, our GREAT-IGEV obtains an even lower occluded EPE of 1.51 and eliminates the need for yielding an additional occlusion mask. Meanwhile, among all published methods, our GREAT-IGEV ranks first on both KITTI 2015 \cite{menze2015object} and ETH3D \cite{schops2017multi} leaderboards and achieves second on Middlebury \cite{scharstein2014high} benchmark. Our main contributions can be summarized as follows:
\begin{itemize}
    \item We propose a universal framework that can be integrated into existing iterative stereo-matching methods to improve the performance in ill-posed regions.
    \item We introduce Spatial (SA), Matching (MA), and Volume (VA) Attentions, designed to mitigate ambiguities in ill-posed regions with global context information.
    \item Our method outperforms existing published methods on public leaderboards such as SceneFlow, KITTI, ETH3D, and Middlebury, with especially significant improvements in ill-posed regions.
\end{itemize}

\section{Related Work}

\begin{figure*}[tp]
    \centering
    \includegraphics[width=0.89\linewidth]{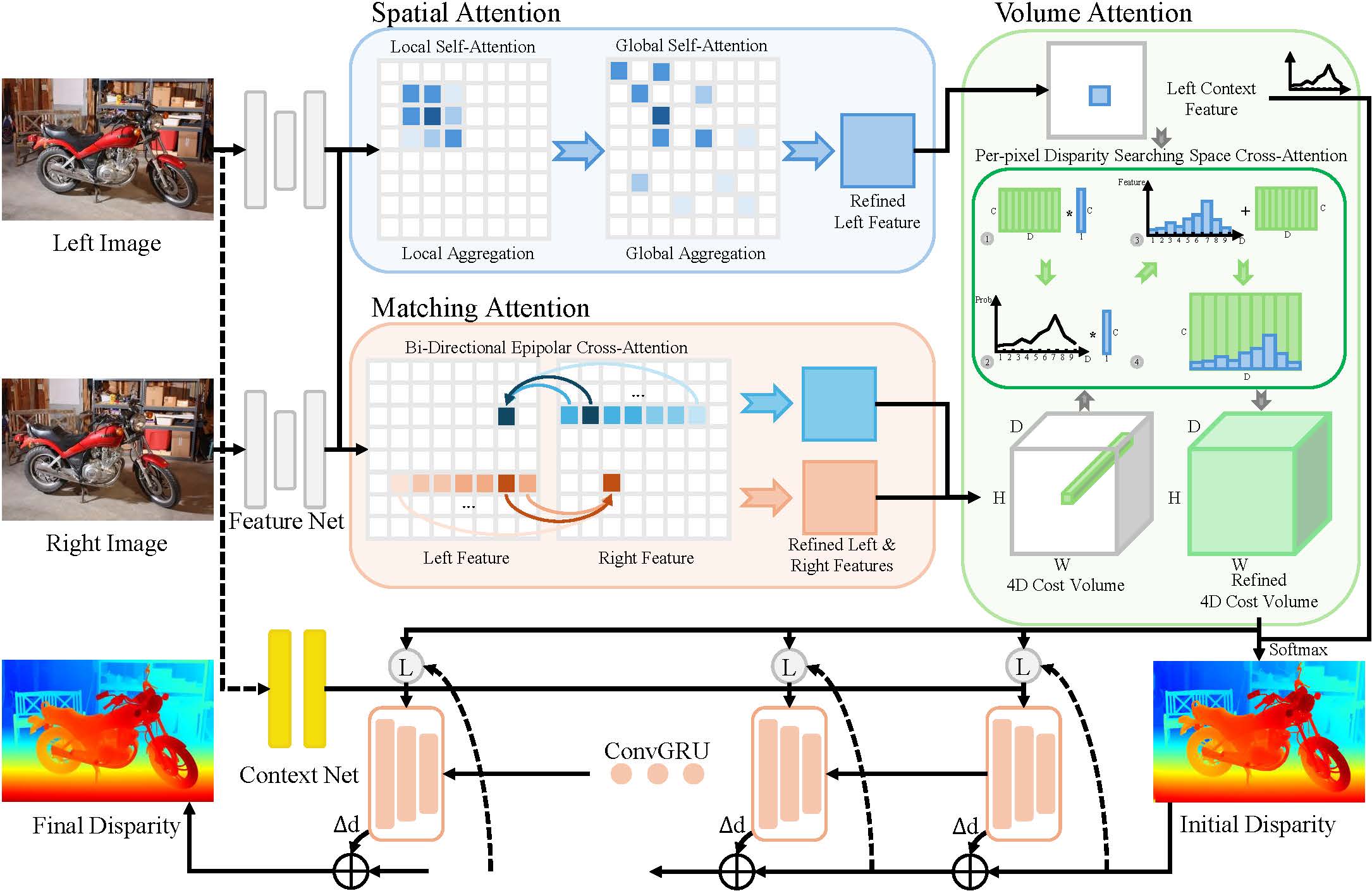}
    \caption{Overview of our proposed GREAT-Stereo (GREAT-IGEV version). The GREAT-Stereo first constructs the cost volume with Matching Attention (MA) which encodes global context along epipolar lines in left and right features. Then, Spatial Attention (SA) extracts global context information from the spatial dimension of left feature and incorporates with Volume Attention (VA) to further involve global geometry in the cost volume which is used to generate an initialized disparity map and participate in iterations of ConvGRU.}
    \label{fig:architecture}
\end{figure*}

\label{sec:related}
\textbf{Aggregation-based \& Iterative Approaches.} Thanks to deep learning, there has been a proliferation of neural network architectures for stereo matching~\cite{chang2018pyramid, cheng2022region, cheng2024coatrsnet, guo2019group, kendall2017end, shen2022pcw, xu2022attention, xu2023accurate, xu2020aanet, zhang2019ga, zhang2024learning, wang2025learning}. Methods such as DispNet \cite{mayer2016large}, GCNet \cite{kendall2017end}, PSMNet \cite{chang2018pyramid}, and ACVNet \cite{xu2022attention}, compute a pixel-wise cost volume and deploy 3D CNN to aggregate it for disparity prediction. However, the high computational demands of 3D CNN limit their efficiency, especially in high-resolution images. To address this, cascaded methods \cite{gu2020cascade, shen2021cfnet, xu2023accurate} have been introduced, which employ a coarse-to-fine strategy to improve efficiency, albeit propagating coarse disparity errors. In response, iterative methods \cite{lipson2021raft, li2022practical, zhao2023high, zeng2023parameterized, xu2023iterative, wang2024selective, xu2025igev++} have been proposed, such as RAFT-Stereo \cite{lipson2021raft} which recurrently updates the disparity using local correlation information retrieved from the cost volume and IGEV-Stereo \cite{xu2023iterative} that integrates aggregation and iterative mechanisms with a lightweight 3D CNN to capture non-local geometric structures. These iterative schemes stand out for balancing computational efficiency and performance.

\textbf{Ill-posed Ambiguities in Stereo Matching.} Despite the observed advancements~\cite{mayer2016large, chang2018pyramid, xu2022attention, lipson2021raft, xu2023iterative}, the ill-posed ambiguities in stereo matching continue to pose a challenge. To combat this issue, researchers have introduced uncertainty-guided approaches \cite{wang2022uncertainty, jing2023uncertainty, liu2024adaptively, liu2024global, zhao2023high}, aimed at improving performance in occluded areas. UGAW-Stereo \cite{jing2023uncertainty} generates an uncertainty map to guide the construction of the cost volume. Similarly, GOAT-Stereo \cite{liu2024global} and ERCNet \cite{liu2024adaptively} use uncertainty maps as error indicators to help identify occlusions and deploy left images to enhance areas characterized by high uncertainty. However, while these uncertainty-guided techniques provide improvements, they do not fully uncover the underlying causes of matching ambiguities in ill-posed regions. GMA-Flow \cite{jiang2021learning} is one of the pioneers that explore the principles behind these ambiguities, noting that global motion can be utilized to propagate information from non-occluded to occluded regions, thereby supporting more reliable predictions in such areas. Analogously, IGEV-Stereo \cite{xu2023iterative} and Selective-Stereo \cite{wang2024selective} suggest that leveraging non-local geometric cues and varying receptive fields can yield robust results for textureless and repetitive texture regions. On top of these findings, our proposed GREAT-Stereo framework extracts global context information from both spatial dimension and epipolar lines to construct a more robust cost volume for stereo matching, which has achieved outstanding performance in tackling ill-posed regions, such as occluded, textureless, and repetitive texture areas.

\section{Method}

\label{sec:method}
In this section, we introduce the GREAT-Stereo, a universal framework designed to improve the performance of various iterative stereo-matching methods in ill-posed regions. We present GREAT-IGEV (Fig. \ref{fig:architecture}) as an example and focus on illustrating its key components.

\subsection{Framework Outline}
\label{sec:outline}
For fair comparisons, GREAT-IGEV maintains consistency with IGEV-Stereo \cite{xu2023iterative} by employing MobileNetV2 \cite{sandler2018mobilenetv2} for the extraction of multi-scale left and right features $ \mathbf{f}_{l(r),i} \in \mathbb{R}^{C_i \times \frac{H}{i} \times \frac{W}{i}} $ ($ i=4, 8, 16, 32 $ and $ C_i $ for feature channels) from stereo images $ \mathbf{I}_{l(r)} \in \mathbb{R}^{3 \times H \times W} $. Taking the left features $ \mathbf{f}_{l,i}\ (i=4, 8, 16, 32) $ as input, \textbf{Spatial Attention} (SA in Sec. \ref{sec:sa}) aggregates context details in a local-to-global fashion. Concurrently, \textbf{Matching Attention} (MA in Sec. \ref{sec:ma}) captures detailed global context along epipolar lines using the largest-scale left and right features $ \mathbf{f}_{l(r),4} $, facilitating the construction of a concatenation-based cost volume $ \mathbf{C}_{cat} $. This cost volume is refined by \textbf{Volume Attention} (VA in Sec. \ref{sec:va}), which incorporates insights gained from the SA, resulting in a robust cost volume that leads to accurate initial disparity. In line with IGEV-Stereo, the same local correlation sampler \cite{xu2023iterative} is utilized to retrieve information from $ \mathbf{C}_{cat} $, which is then integrated with ConvGRU \cite{lipson2021raft, xu2023iterative} to iteratively predict the final disparity.

\subsection{Spatial Global Context Aggregation}
\label{sec:sa}
To facilitate the propagation of geometric structures in occlusions, the \textbf{Spatial Attention} (SA) captures richer contextual details from multi-scale left features $ \mathbf{f}_{l,i}\ (i=4, 8, 16, 32) $ in a local-to-global fashion by leveraging attention mechanism \cite{vaswani2017attention, dosovitskiy2020vit}. This is accomplished through two attention mechanisms applied to each scale individually: Local Self-Attention (LSA) originated from Vision-Outlooker \cite{yuan2022volo} and Global Self-Attention (GSA) derived from Swin-Transformer \cite{liu2021swin}, as depicted in Fig. \ref{fig:architecture}.

\textbf{Local Self-Attention.} With inputs $ \mathbf{f}_{l,i}\ (i=4, 8, 16, 32) $, LSA compensates for local geometry and generates multi-scale local geometric left features $ \mathbf{f}_{l,i}^{local} \in \mathbb{R}^{C_i \times \frac{H}{i} \times \frac{W}{i}} \ (i=4, 8, 16, 32) $ using Outlook Attention \cite{yuan2022volo}, formulated as:
\begin{equation}
    \begin{aligned}
        \mathbf{f}_{l,i}^{local} &= \mathrm{OutlookAttn}(\mathrm{LN}(\mathbf{f}_{l,i})) + \mathbf{f}_{l,i} \\
        \mathbf{f}_{l,i}^{local} &= \mathrm{MLP}(\mathrm{LN}(\mathbf{f}_{l,i}^{local})) + \mathbf{f}_{l,i}^{local}
    \end{aligned}
    \label{func:volo}
\end{equation}
where $ \mathrm{LN} $ stands for LayerNorm \cite{liu2021rethinking} and $ \mathrm{MLP} $ refers to a multilayer perceptron. Outlook Attention employs a $ K \times K $ window centered at each spatial location $ (m, n) $ within $ \mathbf{f}_{l,i} $ and involves three linear projections: $ to\_attn $ computes the attention weights $ \mathbf{W}_{A}^{outlook} \in \mathbb{R}^{K^4} $ from vector $ \mathbf{v}_{l,i} \in \mathbb{R}^{C_i} $ at location $ (m, n) $; $ to\_value $ generates value features $ \mathbf{W}_{V}^{outlook} \in \mathbb{R}^{C_i \times K^2} $ from assigned windows; and $ to\_out $ aggregates final vector $ \mathbf{v}_{l,i}^{local} \in \mathbb{R}^{C_i} $ at location $ (m, n) $ by multiplying the softmax of $ \mathbf{W}_{A}^{outlook} $ and $ \mathbf{W}_{V}^{outlook} $ as follows:
\begin{equation}
    \begin{aligned}
        \mathbf{v}_{l,i}^{local} &= \sum_{k=1}^{K^2} \mathrm{Softmax(\mathbf{W}_{A}^{outlook})}\ \cdot\ \mathbf{W}_{V}^{outlook} \\
        \mathbf{v}_{l,i}^{local} &= \mathrm{to\_out}(\mathbf{v}_{l,i}^{local})
    \end{aligned}
    \label{func:outlook_attention}
\end{equation}

\textbf{Global Self-Attention.} Utilizing $ \mathbf{f}_{l, i}^{local}\ (i=4, 8, 16, 32) $ from LSA, GSA encodes global geometry across the entire spatial dimension, producing multi-scale global geometric left features $ \mathbf{f}_{l,i}^{spatial} \in \mathbb{R}^{C_i \times \frac{H}{i} \times \frac{W}{i}}\ (i=4, 8, 16, 32) $. To address the high computational cost of full spatial dimensions $ (\mathbf{H} \times \mathbf{W}) $, Window Attention from Swin-Transformer \cite{liu2021swin} has been implemented as follows:
\begin{equation}
    \begin{aligned}
        \mathbf{f}_{l,i}^{local} &= \mathrm{PE}(\mathbf{f}_{l,i}^{local}) + \mathbf{f}_{l,i}^{local} \\
        \mathbf{f}_{l,i}^{spatial} &= \mathrm{WSA}(\mathrm{LN}(\mathbf{f}_{l,i}^{local})) + \mathbf{f}_{l,i}^{local} \\
        \mathbf{f}_{l,i}^{spatial} &= \mathrm{MLP}(\mathrm{LN}(\mathbf{f}_{l,i}^{spatial})) + \mathbf{f}_{l,i}^{spatial} \\
        \mathbf{f}_{l,i}^{spatial} &= \mathrm{SWSA}(\mathrm{LN}(\mathbf{f}_{l,i}^{spatial})) + \mathbf{f}_{l,i}^{spatial} \\
        \mathbf{f}_{l,i}^{spatial} &= \mathrm{MLP}(\mathrm{LN}(\mathbf{f}_{l,i}^{spatial})) + \mathbf{f}_{l,i}^{spatial}
    \end{aligned}
    \label{func:window_attention}
\end{equation}
where $ \mathrm{PE} $ denotes positional embedding \cite{vaswani2017attention, dosovitskiy2020vit}, while $ \mathrm{WSA} $ and $ \mathrm{SWSA} $ are Window Self-Attention and Shifted-Window Self-Attention, respectively. $ \mathrm{WSA} $ divides the whole spatial dimension into $ k $ windows of size $ \frac{H}{k} \times \frac{W}{k} $, reducing memory cost by calculating attention within each window. To ensure completed global aggregation, $ \mathrm{SWSA} $ computes the attention map inside shifted windows by rolling it from left-to-right and up-to-down with strides of $ \frac{W}{2k} $ and $ \frac{H}{2k} $, respectively.

\subsection{Bi-Directional Epipolar Global Refinement}
\label{sec:ma}
To mitigate the matching ambiguities in textureless and repetitive texture regions, \textbf{Matching Attention} (MA) captures global context along epipolar lines using both largest-scale of left $ \mathbf{f}_{l,4} $ and right $ \mathbf{f}_{r,4} $ features. This is achieved via Bi-Directional Epipolar Cross-Attention (BECA), which utilizes one-dimensional attention, as illustrated in Fig. \ref{fig:architecture}.

\textbf{Mitigation of Imperfective Rectification.} The BECA enhances global matching along epipolar lines using both left $ \mathbf{f}_{l,4} $ and right $ \mathbf{f}_{r,4} $ features, increasing sensitivity to rectification imperfections. Consequently, non-parallel epipolar lines may introduce additional noise post-BECA. To mitigate this, a streamlined pipeline encodes information from varying receptive fields around the epipolar lines using the Contextual Spatial Attention Module in Selective-Stereo \cite{wang2024selective}. Specifically, two CNNs with small (kernel = 3) and large (kernel = 5) receptive fields aggregate different frequency information from $ \mathbf{f}_{l,4} $ and $ \mathbf{f}_{r,4} $. Then an attention map from \cite{wang2024selective} fuses these features ($ \mathbf{f}_{l(r),4}^{fused} \in \mathbb{R}^{C_4 \times \frac{H}{4} \times \frac{W}{4}}$), encoding more local geometry around the epipolar lines and thereby enhancing robustness to imperfect rectification.

\textbf{Bi-Directional Epipolar Cross Attention.} Given the fused features $ \mathbf{f}_{l,4}^{fused} $ and $ \mathbf{f}_{r,4}^{fused} $ derived from stereo images $ \mathbf{I}_{l} $ and $ \mathbf{I}_{r} $, BECA produces matching left and right features $ (\mathbf{f}_{l(r),4}^{mat} \in \mathbb{R}^{C_4 \times \frac{H}{4} \times \frac{W}{4}}) $ for constructing a robust cost volume. To expedite bi-directional cross-attention from left to right and vice versa, these feature maps are first concatenated along the batch dimension, formulated as follows:
\begin{equation}
    \begin{aligned}
        \mathbf{f}_{l\rightarrow r,4}^{cat} &= <\mathbf{f}_{l,4}^{fused}\ |\ \mathbf{f}_{r,4}^{fused}> \\
        \mathbf{f}_{r\rightarrow l,4}^{cat} &= <\mathbf{f}_{r,4}^{fused}\ |\ \mathbf{f}_{l,4}^{fused}> \\
    \end{aligned}
    \label{func:concatenation}
\end{equation}
where $ <A\ |\ B> $ denotes concatenation between features $ A $ and $ B $, yielding $ \mathbf{f}_{*,4}^{cat} \in \mathbb{R}^{2 \times C_4 \times \frac{H}{4} \times \frac{W}{4}} $ with $ * $ refers to $ l\rightarrow r $ or $ r\rightarrow l $. Subsequently, one-dimensional cross-attention is applied along the width to generate an attention map of shape $ \frac{W}{4} \times \frac{W}{4} $, aggregating global context along epipolar lines, defined as follows:
\begin{equation}
    \begin{aligned}
        \mathbf{f}_{*,4}^{cat} &= \mathrm{LN}(\mathrm{PE}(\mathbf{f}_{*,4}^{cat}) + \mathbf{f}_{*,4}^{cat}) \\
        \mathbf{f}_{l\rightarrow r,4}^{cat} &= \mathrm{CrossAttn}(\mathbf{f}_{l\rightarrow r,4}^{cat}, \mathbf{f}_{r\rightarrow l,4}^{cat}) + \mathbf{f}_{l\rightarrow r,4}^{cat} \\
        \mathbf{f}_{l\rightarrow r,4}^{cat} &= \mathrm{MLP}(\mathrm{LN}(\mathbf{f}_{l\rightarrow r,4}^{cat})) + \mathbf{f}_{l\rightarrow r,4}^{cat}
    \end{aligned}
    \label{func:matching_attention}
\end{equation}
Finally, $ \mathbf{f}_{l\rightarrow r,4}^{cat} $ is decomposed into $ \mathbf{f}_{l,4}^{mat} $ and $ \mathbf{f}_{r,4}^{mat} $ along batch axis for the subsequent processing.

\subsection{Cost Volume Construction and Refinement}
\label{sec:va}
To directly introduce global context and geometric structures into cost volume, \textbf{Volume Attention} (VA) excites global context details from specific positions within cost volume by leveraging the spatial features $ \mathbf{f}_{l,i}^{spatial}\ (i=4, 8, 16, 32) $ derived from Spatial Attention (Sec. \ref{sec:sa}).

\textbf{Concatenated Cost Volume.} To accurately identify positions that require global context excitation within the cost volume, it is essential to ensure consistency in feature distribution between $ \mathbf{f}_{l,i}^{spatial} $ and cost volume. Consequently, the cost volume $ \mathbf{C}_{cat} $ is constructed by concatenating matching features $ \mathbf{f}_{l(r),4}^{mat} $ from Matching Attention (Sec. \ref{sec:ma}) along the channel dimension, rather than computing correlations between them, formulated as follows:
\begin{equation}
    \begin{aligned}
        \mathbf{C}_{cat}(\cdot, d, x, y) &= <\mathbf{f}_{l,4}^{mat}(x, y)\ |\ \mathbf{f}_{r,4}^{mat}(x - d, y)>
    \end{aligned}
    \label{func:cost_volume}
\end{equation}
where $ \mathbf{C}_{cat} \in \mathbb{R}^{2C_{4} \times D \times \frac{H}{4} \times \frac{W}{4}} $ denotes the 4D concatenated cost volume, with $ d \in \{0, 1, \dots, D - 1\} $ referring the $ d^{th} $ disparity searching space. We also calculate all-pairs correlations \cite{lipson2021raft} to ensure consistency with IGEV-Stereo.

\textbf{Per-pixel Disparity Searching Space Cross-Attention.} The core technique of VA, Per-pixel Disparity Searching Space Cross-Attention (PDCA), is integrated into each scale of the UNet in \cite{xu2023iterative}, replacing the Feature Attention from \cite{bangunharcana2021correlate}. As shown in Fig. \ref{fig:architecture}, PDCA employs a soft-classification scheme to identify positions in cost volume that require global context enhancement through per-pixel cross-attention, defined as follows:
\begin{equation}
    \begin{aligned}
        \mathbf{A}_{m,n} &= \mathrm{Softmax}(\mathbf{M}_{m,n}^{\mathbf{C}}\ \cdot\ \mathbf{v}_{m,n}^{\mathbf{f}}) \\
        \mathbf{M}_{m,n}^{\mathbf{C}} &= \mathrm{MLP}(\mathbf{A}_{m,n}\ \cdot\ \mathbf{v}_{m,n}^{\mathbf{f}}) + \mathbf{M}_{m,n}^{\mathbf{C}}
    \end{aligned}
    \label{func:volume_attention}
\end{equation}
where $ \mathbf{M}_{m,n}^{\mathbf{C}} \in \mathbb{R}^{C_4 \times D} $ represents the disparity searching space matrix at pixel location $ (m,n) $ in $ \mathbf{C}_{cat} $, $ \mathbf{v}_{m,n}^{\mathbf{f}} \in \mathbb{R}^{C_4} $ is the spatial feature vector at the same location in $ \mathbf{f}_{l,i}^{spatial} $, and $ \mathbf{A}_{m,n} \in \mathbb{R}^{D} $ denotes the attention map specifying the enhancement proportion in the disparity searching space. This formulation enables PDCA to directly excite global context in specific positions within the cost volume by referring to spatial features, instead of solely filtering the distribution like in Feature Attention \cite{bangunharcana2021correlate}.

\subsection{Loss Function}
\label{sec:loss}
Aligned with IGEV-Stereo \cite{xu2023iterative}, the initial disparity $ \mathbf{d}_0 $ regressed from $ \mathbf{C}_{cat} $ is supervised by Smooth L1 loss \cite{chang2018pyramid}, while an L1 loss with exponentially increasing weights \cite{lipson2021raft} is applied to all updated disparities $ \{\mathbf{d}_{i}\}_{i=1}^{N} $. The total loss is defined as follows:
\begin{equation}
    \begin{aligned}
        \mathcal{L}_{init} &= \mathrm{Smooth}_{\mathrm{L_1}}(\mathbf{d}_0 - \mathbf{d}_{gt}) \\
        \mathcal{L}_{stereo} &= \mathcal{L}_{init} + \sum\nolimits_{i=1}^{N}\gamma^{N-i}||\mathbf{d}_{i} - \mathbf{d}_{gt}||_1
    \end{aligned}
\end{equation}
where $ \mathbf{d}_{gt} $ denotes the ground truth disparity and $ \gamma = 0.9 $.

\section{Experiments}
\label{sec:experiments}
In this section, we conduct extensive experiments and ablation studies to validate the effectiveness and universality of our proposed GREAT-Stereo framework.

\begin{figure*}[tp]
    \centering
    \includegraphics[width=0.85\linewidth]{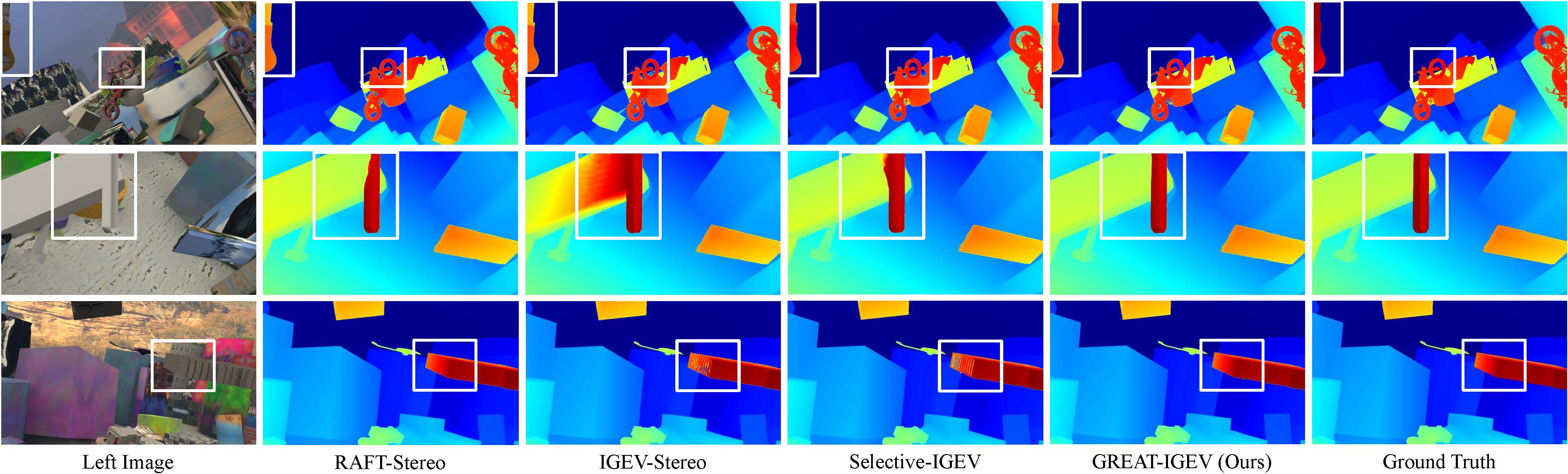}
    \caption{Qualitative results on the Scene Flow test set. Our GREAT-IGEV outperforms other iterative methods in occluded (\textbf{Row 1}), textureless (\textbf{Row 2}), and repetitive texture (\textbf{Row 3}) regions.}
    \label{fig:sceneflow_vis}
\end{figure*}
\begin{table*}[t]
\centering
\resizebox{1.95\columnwidth}{!}{%

\begin{tabular}{l|cccc|cccccc|c}
\toprule
\multirow{2}{*}{Model} & \multirow{2}{*}{Corr} & \multirow{2}{*}{VA} & \multirow{2}{*}{SA} & \multirow{2}{*}{MA} & \multicolumn{2}{c}{All} & \multicolumn{2}{c}{Non-Occ} & \multicolumn{2}{c|}{Occ} & \multirow{2}{*}{Param (M)} \\
 & & & & & EPE (px) & D3 (\%) & EPE (px) & D3 (\%) & EPE (px) & D3 (\%) & \\
\hline
Baseline (IGEV-Stereo) & \checkmark & & & & 0.479 & 2.476 & 0.194 & 0.633 & 1.649 & 10.980 & 12.60 \\
\hline
$ \mathbf{C}_{corr} $ + Volume & \checkmark & \checkmark & & & 0.469 & 2.443 & 0.187 & 0.621 & 1.623 & 10.914 & 12.91 \\
$ \mathbf{C}_{cat} $ + Volume & & \checkmark & & & 0.466 & 2.437 & 0.183 & 0.612 & 1.616 & 10.860 & 12.91 \\
$ \mathbf{C}_{cat} $ + Volume + Spatial & & \checkmark & \checkmark & & 0.457 & 2.418 & 0.181 & 0.623 & 1.587 & 10.698 & 13.62 \\
$ \mathbf{C}_{cat} $ + Matching & & & & \checkmark & 0.430 & 2.229 & 0.153 & \textbf{0.483} & 1.558 & 10.283 & 13.40 \\
Full Model (GREAT-IGEV) & & \checkmark & \checkmark & \checkmark & \textbf{0.413} & \textbf{2.201} & \textbf{0.135} & 0.486 & \textbf{1.514} & \textbf{10.115} & 14.44 \\
\bottomrule
\end{tabular}

}
\caption{
Ablation study of the effectiveness of the proposed modules on the Scene Flow test set (\textbf{bold}: best). Corr denotes the correlation-based cost volume, Volume (VA) refers to Volume Attention, Spatial (SA) represents Spatial Attention, and Matching (MA) indicates Matching Attention. EPE means the End-Point Error to describe the overall errors. D3 signifies the proportion of pixels with errors exceeding 3 px. All, Non-Occ, and Occ respectively denote the entire, non-occluded, and occluded regions.
}
\vspace{-0.05in}
\label{tab:ablation_of_modules}
\end{table*}

\begin{table}[t]
\centering
\resizebox{0.88\columnwidth}{!}{%

\begin{tabular}{lcccc}
\toprule
Model & All & Non-Occ & Occ & Param (M) \\
\hline
RAFT-Stereo \cite{lipson2021raft} & 0.551 & 0.214 & 1.938 & 11.12 \\
GREAT-RAFT & \textbf{0.488} & \textbf{0.183} & \textbf{1.747} & 12.40 \\
\hline
IGEV-Stereo \cite{xu2023iterative} & 0.479 & 0.194 & 1.649 & 12.60 \\
GREAT-IGEV & \textbf{0.413} & \textbf{0.135} & \textbf{1.514} & 14.44 \\
\hline
Selective-IGEV \cite{wang2024selective} & 0.453 & 0.173 & 1.574 & 13.14 \\
GREAT-Selective & \textbf{0.419} & \textbf{0.146} & \textbf{1.524} & 14.98 \\
\bottomrule
\end{tabular}

}
\caption{
Ablation study of the cross-model transferability of the proposed framework (\textbf{bold}: best). Results report the End-Point Error (EPE) in corresponding regions on the Scene Flow test set.
}
\label{tab:ablation_of_transferability}
\end{table}

\subsection{Datasets}
\label{sec:dataset}
\textbf{Scene Flow} \cite{mayer2016large}, a synthetic dataset, includes 35454 training and 4370 testing pairs with dense disparity maps. We utilize the finalpass version for its more realistic and challenging characteristics over the cleanpass version. \textbf{KITTI 2012} \cite{geiger2012we} and \textbf{KITTI 2015} \cite{menze2015object} provide real-world driving scene data. KITTI 2012 offers 194 training and 195 testing pairs, and KITTI 2015 supplies 200 training and 200 testing pairs. \textbf{ETH3D} \cite{schops2017multi} includes 27 training and 20 testing gray-scale stereo pairs for both indoor and outdoor scenarios. \textbf{Middlebury} \cite{scharstein2014high} known as high-resolution indoor scenes, comprises 15 training and 15 testing pairs.

\subsection{Implementation Details}
\label{sec:implementation}
We implement GREAT-Stereo with PyTorch and train the models on NVIDIA RTX 3090 and 4090 GPUs. For all experiments, we use the AdamW \cite{loshchilov2017decoupled} optimizer with gradient clipping set to [-1, 1] and the one-cycle \cite{smith2019super} learning rate schedule with a learning rate of 2e-4. Pretraining is conducted on the Scene Flow dataset for 200k steps with a batch size of 8. During pretraining, the image crops are 320 x 720 for models based on RAFT-Stereo \cite{lipson2021raft} and 320 x 736 for models based on IGEV-Stereo \cite{xu2023iterative}, and the number of iterations is set to 22. For the number of attention blocks, each scale applies only one Spatial Attention and one Volume Attention block. Concurrently, the largest scale utilizes four Matching Attention blocks.

\subsection{Ablation Study}
\label{sec:ablation_study}
In this section, we evaluate multiple variants of our framework to verify the proposed modules in several aspects. All results use 32 update iterations.

\textbf{Effectiveness of Proposed Modules.} To assess the proposed framework, we take IGEV-Stereo \cite{xu2023iterative} as the baseline, integrating Spatial (SA), Matching (MA), and Volume (VA) Attentions. As shown in Tab. \ref{tab:ablation_of_modules}, VA universally improves accuracy, even without global context aggregation via SA and MA. This suggests that proactively exciting geometric details by spatial features creates a more robust cost volume. In Tab. \ref{tab:ablation_of_modules}, SA effectively propagates non-occluded information to occluded regions by incorporating global context, thereby enhancing occlusion handling. Additionally, MA enriches global context along epipolar lines, aiding in the disambiguation of textureless or repetitive texture areas (see Fig. \ref{fig:sceneflow_vis}), optimizing non-occluded regions. Consequently, SA and MA enhance the performance of VA, highlighting the effectiveness of the proposed framework in addressing ill-posed issues by enriching the global context.

\textbf{Cross-Model Transferability of Proposed Modules.} To evaluate the universality of the proposed GREAT-Stereo framework, we integrate SA, MA, and VA into three baselines: RAFT-Stereo \cite{lipson2021raft}, IGEV-Stereo \cite{xu2023iterative}, and Selective-IGEV \cite{wang2024selective}. The resulting models are denoted as GREAT-RAFT, GREAT-IGEV, and GREAT-Selective. Particularly, RAFT-Stereo includes only all-pairs correlation and lacks multi-scale spatial features, thus, only MA is integrated to ensure minimal modification and fair comparison. As shown in Tab. \ref{tab:ablation_of_transferability}, integrating these attention mechanisms significantly improves EPE metrics across all regions, including ill-posed areas, on the Scene Flow test set. Notably, incorporating only MA optimizes the EPE of RAFT-Stereo in occlusions by 9.9\%. When using the full framework, GREAT-IGEV and GREAT-Selective reduce EPE in occlusions by 8.2\% and 3.2\%, respectively. As illustrated in Fig. \ref{fig:performance_disp_vis} and \ref{fig:sceneflow_vis}, the integration of SA, MA, and VA significantly improves performance in textureless and repetitive texture regions, enabling GREAT-IGEV and GREAT-Selective to achieve even greater optimizations of 30.4\% and 15.6\% in non-occluded EPE, respectively. In summary, our framework can be seamlessly integrated into various stereo matching methods, significantly enhancing their performance in ill-posed regions.

\begin{table}[t]
\centering
\resizebox{0.88\columnwidth}{!}{%

\begin{tabular}{lcccccc}
\toprule
\multirow{2}{*}{Model} & \multicolumn{6}{c}{Number of Iterations} \\
\cline{2-7}
& 1 & 2 & 4 & 8 & 16 & 32 \\
\hline
RAFT-Stereo \cite{lipson2021raft} & 2.053 & 1.134 & 0.754 & 0.596 & 0.554 & 0.551 \\
GREAT-RAFT & \textbf{1.646} & \textbf{0.950} & \textbf{0.664} & \textbf{0.530} & \textbf{0.492} & \textbf{0.488} \\
\hline
IGEV-Stereo \cite{xu2023iterative} & 0.669 & 0.623 & 0.557 & 0.504 & 0.483 & 0.479 \\
GREAT-IGEV & \textbf{0.572} & \textbf{0.535} & \textbf{0.483} & \textbf{0.439} & \textbf{0.417} & \textbf{0.413} \\
\bottomrule
\end{tabular}
}
\caption{
Ablation study of the number of iterations (\textbf{bold}: best). Results report the EPE on the Scene Flow test set.
}
\label{tab:ablation_of_the_number_of_iterations}
\end{table}

\begin{table*}[t]
\centering
\resizebox{1.95\columnwidth}{!}{%

\begin{tabular}{lcccccccccc}
\toprule
Metrics & GANet \cite{zhang2019ga} & LEA-Stereo \cite{cheng2020hierarchical} & ACVNet \cite{xu2022attention} & GOAT-Stereo \cite{liu2024global} & RAFT-Stereo \cite{lipson2021raft} & IGEV-Stereo \cite{xu2023iterative} & Selective-IGEV \cite{wang2024selective} & GREAT-IGEV (Ours) \\
\hline
EPE-Non-Occ (px) & - & - & - & - & 0.21 & 0.19 & 0.17 & \textbf{0.14} \\
EPE-Occ (px) & - & - & - & 1.53 & 1.94 & 1.65 & 1.57 & \textbf{1.51} \\
EPE-All (px) & 0.84 & 0.78 & 0.48 & 0.47 & 0.55 & 0.48 & 0.45 & \textbf{0.41} \\
\bottomrule
\end{tabular}

}
\caption{
Quantitative evaluation on Scene Flow test set (\textbf{bold}: best).
}
\label{tab:sceneflow_results}
\end{table*}

\begin{table*}[t]
\centering
\resizebox{1.95\columnwidth}{!}{%

\begin{tabular}{l|cccc|cccc|c}
\toprule
\multirow{2}{*}{Model} & \multicolumn{4}{c|}{KITTI 2012} & \multicolumn{4}{c|}{KITTI 2015} & \multirow{2}{*}{Param (M)} \\
& 2-noc & 2-all & 3-noc & 3-all & D1-all & D1-fg & Noc-D1-all & Noc-D1-fg & \\
\hline
LEA-Stereo \cite{cheng2020hierarchical} & 1.90 & 2.39 & 1.13 & 1.45 & 1.65 & 2.91 & 1.51 & 2.65 & 1.80 \\
ACVNet \cite{xu2022attention} & 1.83 & 2.35 & 1.13 & 1.47 & 1.65 & 3.07 & 1.52 & 2.84 & 6.20 \\
ViTAStereo \cite{liu2024playing} & \textbf{1.46} & \textbf{1.80} & \textbf{0.93} & \textbf{1.16} & \underline{1.50} & 2.99 & 1.41 & 2.90 & - \\
\rowcolor{gray!30!white}
IGEV-Stereo \cite{xu2023iterative} & 1.71 & 2.17 & 1.12 & 1.44 & 1.59 & 2.67 & 1.49 & 2.62 & 12.60 \\
\rowcolor{yellow!15!white}
Selective-IGEV \cite{wang2024selective} & 1.59 & 2.05 & 1.07 & 1.38 & 1.55 & 2.61 & 1.44 & 2.55 & 13.14 \\
IGEV++ \cite{xu2025igev++} & 1.56 & 2.03 & 1.04 & 1.36 & 1.51 & \textbf{2.54} & 1.42 & \underline{2.54} & 14.53 \\
\hline
\rowcolor{gray!30!white}
GREAT-IGEV (Ours) & 1.51 & 2.00 & 1.02 & 1.37 & \underline{1.50} & \underline{2.59} & \textbf{1.37} & \textbf{2.51} & 14.44 \\
\rowcolor{yellow!15!white}
GREAT-Selective (Ours) & \underline{1.48} & \underline{1.94} & \underline{1.00} & \underline{1.31} & \textbf{1.49} & 2.62 & \underline{1.40} & 2.60 & 14.98 \\
\bottomrule
\end{tabular}

}
\caption{
Quantitative evaluation on KITTI 2012 and KITTI 2015 benchmarks (\textbf{bold}: best; \underline{underline}: second best).
}
\label{tab:kitti_results}
\end{table*}

\begin{table}[t]
\centering
\resizebox{0.88\columnwidth}{!}{%

\begin{tabular}{lcccc}
\toprule
\multirow{2}{*}{Model} & \multirow{2}{*}{KITTI 2015} & \multicolumn{2}{c}{Middlebury} & \multirow{2}{*}{ETH3D} \\
& & half & quater & \\
\hline
HD3 \cite{yin2019hierarchical} & 26.5 & 37.9 & 20.3 & 54.2 \\
GwCNet \cite{guo2019group} & 22.7 & 34.2 & 18.1 & 30.1 \\
DSMNet \cite{zhang2020domain} & 6.5 & 13.8 & 8.1 & 6.2 \\
\hline
\rowcolor{gray!15!white}
RAFT-Stereo \cite{lipson2021raft} & \textbf{5.8} & 10.0 & 6.7 & \underline{3.2} \\
\rowcolor{gray!30!white}
IGEV-Stereo \cite{xu2023iterative} & 6.0 & 9.5 & 6.2 & 3.6 \\
\rowcolor{yellow!15!white}
Selective-IGEV \cite{wang2024selective} & 6.1 & 9.2 & 6.6 & 5.4 \\
\hline
\rowcolor{gray!15!white}
GREAT-RAFT (Ours) & \textbf{5.8} & 10.2 & 7.0 & \textbf{2.8} \\
\rowcolor{gray!30!white}
GREAT-IGEV (Ours) & \underline{5.9} & \underline{8.6} & \underline{5.1} & 3.8 \\
\rowcolor{yellow!15!white}
GREAT-Selective (Ours) & 6.2 & \textbf{8.1} & \textbf{4.6} & 3.7 \\
\bottomrule
\end{tabular}

}
\caption{
Zero-shot generalization of the proposed framework (\textbf{bold}: best; \underline{underline}: second best). All models are trained on Scene Flow. The 3-pixel error rate is used for KITTI 2015, 2-pixel error rate for Middlebury, and 1-pixel error rate for ETH3D.
}
\label{tab:zero_shot_generalization}
\end{table}

\textbf{Number of Iterations.} Our GREAT-Stereo achieves comparable or better performance with fewer iterations. As shown in Tab. \ref{tab:ablation_of_the_number_of_iterations}, GREAT-IGEV matches the performance of IGEV-Stereo \cite{xu2023iterative} with only 4 iterations. This shows our framework optimizes ill-posed regions and boosts iterative methods' overall performance via a global-context-enhanced robust cost volume, thereby obtaining comparable performance with only a quarter of the iterations. 

\subsection{Zero-shot Generalization} Due to scarce large real-world training data, zero-shot generalization of stereo models is critical. To assess this and verify that baseline improvements do not stem from overfitting, we train the GREAT-RAFT/IGEV/Selective on synthetic Scene Flow \cite{lipson2021raft} dataset, then test them on real-world KITTI 2015 \cite{menze2015object}, Middlebury \cite{scharstein2014high}, and ETH3D \cite{schops2017multi} training sets without fine-tuning. As shown in Tab. \ref{tab:zero_shot_generalization}, our framework matches or outperforms in the same zero-shot setting. 

\subsection{Comparisons with State-of-the-art} All fine-tuned models are initialized with the pretrained Scene Flow model. Each target dataset uses distinct fine-tuning strategies. For validations, GREAT-RAFT is benchmarked against RAFT-Stereo \cite{lipson2021raft}, GREAT-IGEV against IGEV-Stereo \cite{xu2023iterative}, and GREAT-Selective against Selective-IGEV \cite{wang2024selective}, using each respective model as the baseline.

\textbf{Scene Flow.} As shown in Tab. \ref{tab:ablation_of_transferability} and \ref{tab:sceneflow_results}, our GREAT-IGEV model achieves a state-of-the-art overall EPE of 0.413 on the Scene Flow test set, representing a 13.8\% optimization over IGEV-Stereo \cite{xu2023iterative} and a significant 47.4\% better than LEA-Stereo \cite{cheng2020hierarchical}. Additionally, GREAT-Selective attains a comparable overall EPE of 0.419, surpassing Selective-IGEV \cite{wang2024selective} by 7.5\%. To assess the performance of our model in ambiguous regions, we partition the Scene Flow test set into Non-Occluded and Occluded areas using a ground truth occlusion mask. As shown in Tab. \ref{tab:sceneflow_results}, GREAT-IGEV outperforms IGEV-Stereo \cite{xu2023iterative} by 8.2\% on occluded EPE and achieves even lower occluded EPE of 1.51 compared to GOAT-Stereo \cite{liu2024global} (1.53) which requires to predict an additional occlusion mask. As indicated in Tab. \ref{tab:ablation_of_modules} and Fig. \ref{fig:sceneflow_vis}, GREAT-IGEV excels in textureless and repetitive texture areas, resulting in a 30.4\% reduction on non-occluded EPE compared to IGEV-Stereo.

\begin{figure}[tp]
    \centering
    \includegraphics[width=\linewidth]{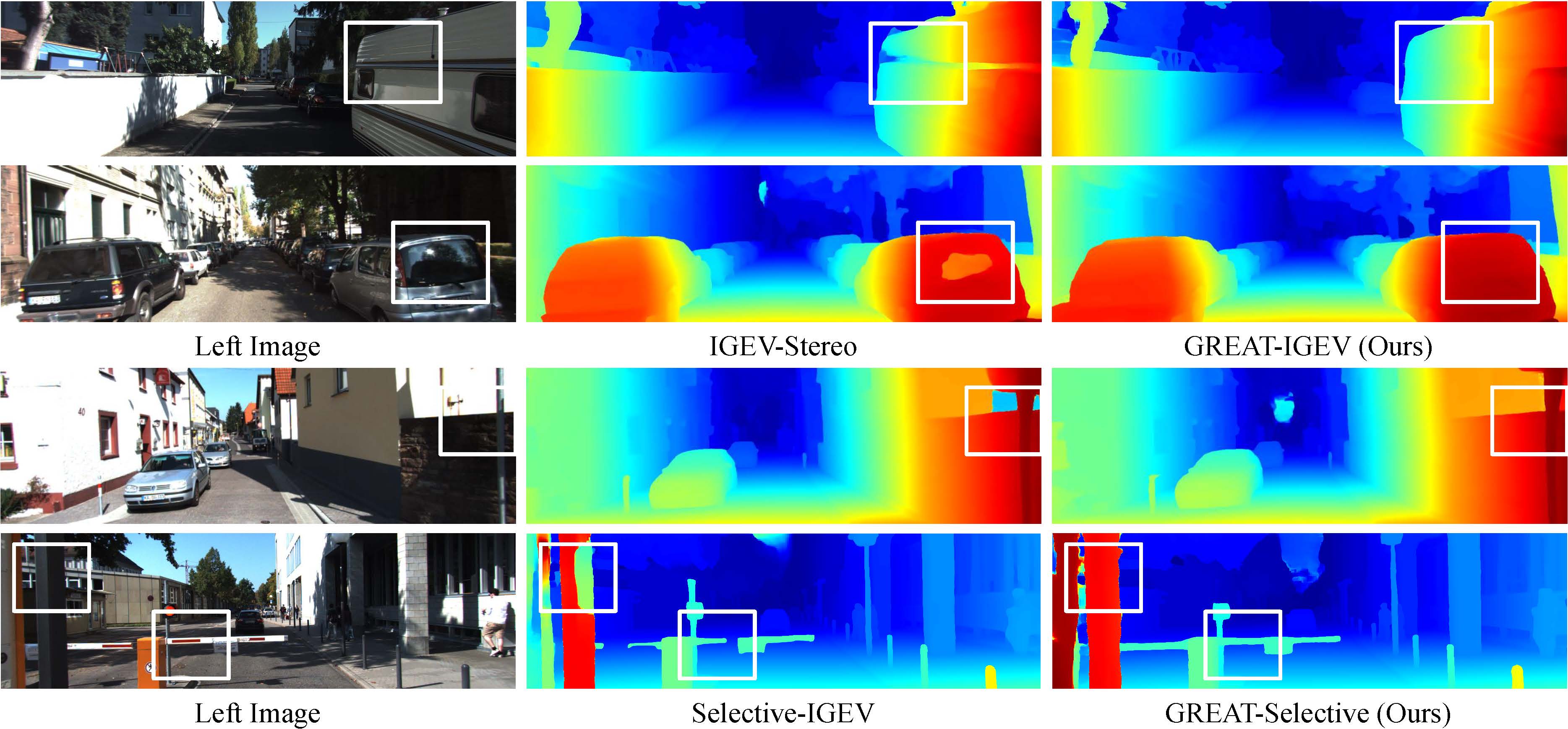}
    \caption{Qualitative results on the KITTI test set. \textbf{Rows 1 \& 2} are the KITTI 2012, and \textbf{Rows 3 \& 4} are the KITTI 2015.}
    \label{fig:kitti_vis}
\end{figure}
\begin{figure*}[tp]
    \centering
    \includegraphics[width=0.84\linewidth]{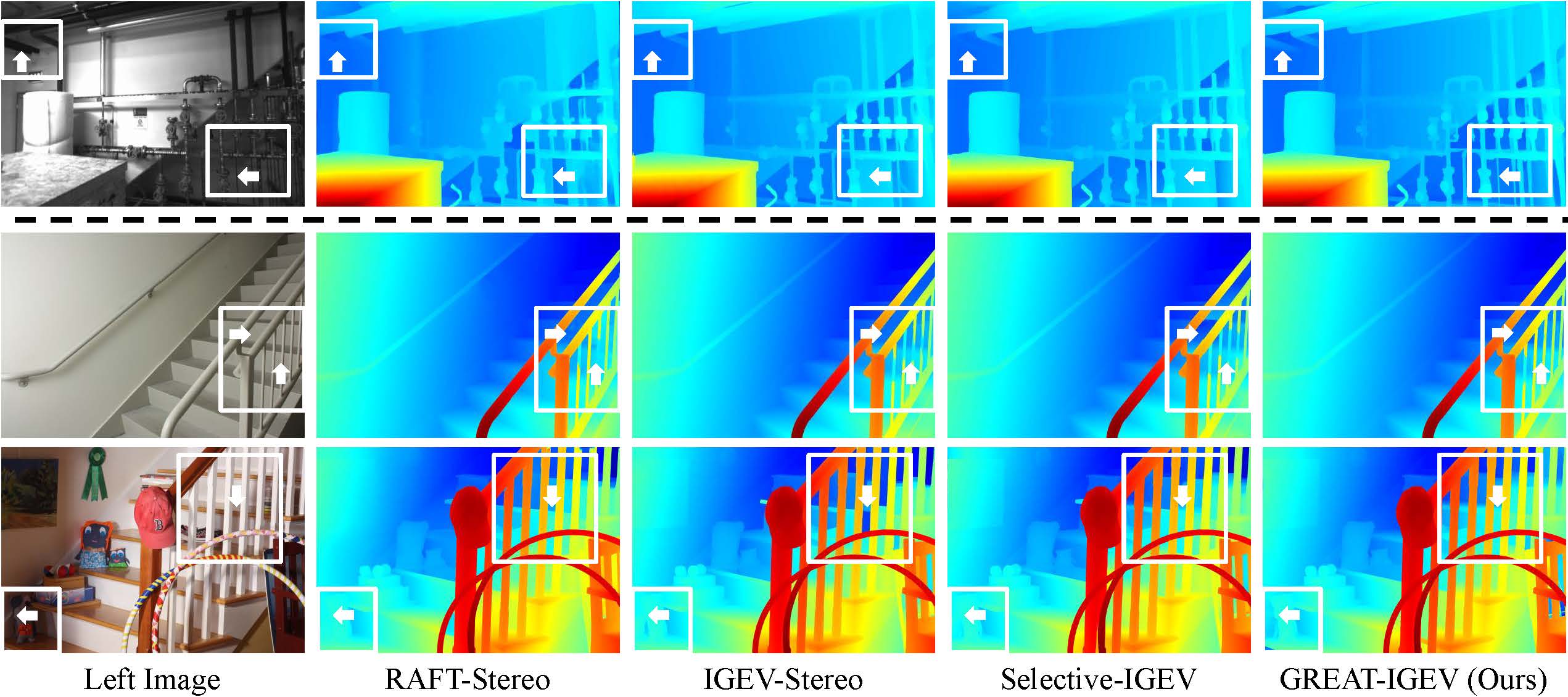}
    \caption{Qualitative results on the ETH3D (\textbf{Row 1}) and Middlebury (\textbf{Row 2 \& 3}) test sets. Our GREAT-IGEV produces clearer and more consistent geometric structures in large textureless and repetitive texture regions, compared to IGEV-Stereo.}
    \label{fig:eth3d_middlebury_vis}
\end{figure*}
\begin{table}[t]
\centering
\resizebox{0.90\columnwidth}{!}{%

\begin{tabular}{l|cc|cc}
\toprule
\multirow{2}{*}{Model} & \multicolumn{2}{c|}{ETH3D} & \multicolumn{2}{c}{Middlebury} \\
& Bad 0.5 & Bad 1.0 & Bad 2.0 & Bad 4.0 \\
\hline
GM-Stereo \cite{xu2023unifying} & 5.94 & 1.83 & 7.14 & 2.96 \\
RAFT-Stereo \cite{lipson2021raft} & 7.04 & 2.44 & 4.74 & 2.75 \\
CRE-Stereo \cite{li2022practical} & 3.58 & \underline{0.98} & 3.71 & 2.04 \\
\rowcolor{gray!30!white}
IGEV-Stereo \cite{xu2023iterative} & 3.52 & 1.12 & 4.83 & 3.33 \\
Selective-IGEV \cite{wang2024selective} & 3.06 & 1.23 & \textbf{2.51} & \textbf{1.36} \\
IGEV++ \cite{xu2025igev++} & \underline{2.98} & 1.14 & 3.23 & 1.82 \\
\rowcolor{gray!30!white}
GREAT-IGEV (Ours) & \textbf{2.18} & \textbf{0.72} & \underline{2.81} & \underline{1.73} \\
\bottomrule
\end{tabular}

}
\caption{
Quantitative evaluation on ETH3D and Middlebury benchmarks (\textbf{bold}: best; \underline{underline}: second best).
}
\label{tab:eth3d_middlebury_results}
\end{table}

\textbf{KITTI.} Our model is first fine-tuned on VKITTI2 \cite{cabon2020virtual} for 60k steps, followed by further fine-tuning on a mixed dataset of KITTI 2012 and 2015 for 50k steps with a batch size of 8. As shown in Tab. \ref{tab:kitti_results}, our models rank first on KITTI 2015 and second on KITTI 2012 among all published methods. On KITTI 2012, our GREAT-Selective surpasses Selective-IGEV \cite{wang2024selective} by 6.9\% on 2-noc metric. On KITTI 2015, our GREAT-IGEV outperforms IGEV-Stereo \cite{xu2023iterative} by 8.1\% on Noc-D1-all metric. As shown in Fig. \ref{fig:kitti_vis}, both GREAT-IGEV and GREAT-Selective demonstrate superior performance over their counterparts in occluded and textureless regions.

\textbf{ETH3D.} Following CRE-Stereo \cite{li2022practical} and GM-Stereo \cite{xu2023unifying}, we use a variety of public stereo datasets for training, with a crop size of 384 x 512 and a batch size of 8. Initially, the Scene Flow pretrained model is fine-tuned on mixed Tartan Air \cite{wang2020tartanair}, CREStereo \cite{li2022practical}, Scene Flow \cite{mayer2016large}, Sintel Stereo \cite{butler2012naturalistic}, InStereo2k \cite{bao2020instereo2k} and ETH3D \cite{schops2017multi} datasets for 300k steps. Then we fine-tune it on the mixed CREStereo \cite{li2022practical}, InStereo2k \cite{bao2020instereo2k} and ETH3D \cite{schops2017multi} datasets for another 100k steps. As shown in Tab. \ref{tab:eth3d_middlebury_results}, our GREAT-IGEV ranks first among all published methods, outperforming IGEV-Stereo \cite{xu2023iterative} by 38.1\% on Bad 0.5 metric. Moreover, GREAT-IGEV exhibits superior performance in occluded and textureless regions, as illustrated in Fig. \ref{fig:eth3d_middlebury_vis}.

\textbf{Middlebury.} In line with CRE-Stereo \cite{li2022practical} and GM-Stereo \cite{xu2023unifying}, we first fine-tune the Scene Flow pretrained model on the mixed Tartan Air \cite{wang2020tartanair}, CREStereo \cite{li2022practical}, Scene Flow \cite{mayer2016large}, Falling Things \cite{tremblay2018falling}, InStereo2k \cite{bao2020instereo2k}, CARLA HR-VS \cite{yang2019hierarchical} and Middlebury \cite{scharstein2014high} datasets using a crop size of 384 x 512 for 200k steps. Subsequently, we fine-tune it on the mixed CREStereo \cite{li2022practical}, Falling Things \cite{tremblay2018falling}, InStereo2k \cite{bao2020instereo2k}, CARLA HR-VS \cite{yang2019hierarchical} and Middlebury \cite{scharstein2014high} datasets using a crop size of 384 x 768 with a batch size of 8 for another 100k steps. As shown in Tab. \ref{tab:eth3d_middlebury_results}, our GREAT-IGEV ranks second among all published methods, outperforming IGEV-Stereo \cite{xu2023iterative} by 41.8\% on Bad 2.0 metric. Moreover, GREAT-IGEV exhibits superior performance in occluded, textureless, and repetitive texture regions, as illustrated in Fig. \ref{fig:eth3d_middlebury_vis}.

\section{Conclusion and Future Work}
\label{sec:conclusion}
We introduce GREAT-Stereo, a novel framework compatible with various iterative stereo-matching methods. Integrating Spatial (SA) and Matching (MA) Attention, the network enhances global context from spatial dimensions and epipolar lines. The combination of SA and MA facilitates Volume Attention (VA) which excites the global context within cost volume for a more robust construction. GREAT-Stereo leads in Scene Flow, KITTI, ETH3D, and Middlebury benchmarks. It demonstrates superior capability in leveraging global context to resolve matching ambiguities in ill-posed regions.

On top of our proposed framework, we plan to investigate two extra challenges in the future: 1) MA incurs intensive computational costs with long epipolar lines; 2) Among the ill-posed areas, the reflective ones pose a suboptimal challenge caused by specular surface lighting conditions rather than lacking global geometric factors, hence requiring an extra module to handle these areas.
\section*{Acknowledgement}
\label{sec:acknowledgement}
{\sloppy The work is supported by a grant from Hong Kong Research Grant Council under GRF 11210622. \par}

{
    \small
    \bibliographystyle{ieeenat_fullname}
    \bibliography{main}
}

\end{document}